\pdfoutput=1

\documentclass[11pt]{article}

\usepackage[]{ACL2023}

\usepackage{times}
\usepackage{latexsym}
\usepackage{tabularx}

\usepackage[T1]{fontenc}

\usepackage[utf8]{inputenc}

\usepackage{microtype}

\usepackage{inconsolata}

\usepackage{todonotes}
\usepackage{multirow}
\usepackage{graphicx}
\usepackage{enumitem}
\usepackage{anyfontsize}
\usepackage{url}
\usepackage{subcaption}
\usepackage{xr}
\usepackage{tabu}
\usepackage{bbold}
\usepackage{booktabs}
\usepackage{xcolor,pifont}
\usepackage{MnSymbol,wasysym}
\usepackage{fontawesome}
\usepackage{xspace}
\usepackage{caption}
\usepackage{array}
\newcolumntype{H}{>{\setbox0=\hbox\bgroup}c<{\egroup}@{}}

\newcommand{\mdma}[1]{{\color{black}{#1}}}

\newcommand{\Skip}[1]{}

\newcommand{\ie}{\textit{i}.\textit{e}.\ }

\newcommand{\secref}[1]{Section~\ref{#1}}
\newcommand{\appref}[1]{Appendix~\ref{#1}}
\newcommand{\apprefsimple}[1]{\ref{#1}}
\newcommand{\figref}[1]{Figure~\ref{#1}}
\newcommand{\tbref}[1]{Table~\ref{#1}}

\newcommand{\dotieconcat}[2]{
  \text{\raisebox{.8ex}{$\smallfrown$}}%
}
\newcommand{\mypar}[1]{\paragraph{#1}}
 
\title{Parameter-Efficient Low-Resource \\ Dialogue State Tracking by Prompt Tuning}

\newcommand\blfootnote[1]{%
  \begingroup
  \renewcommand\thefootnote{}\footnote{#1}%
  \addtocounter{footnote}{-1}%
  \endgroup
}
\author{
    Mingyu Derek Ma$^{\dagger}$\quad
    Jiun-Yu Kao$^{\ddagger}$\quad
    Shuyang Gao$^{\mathsection}$\quad
    Arpit Gupta$^{\ddagger}$\quad \\
    {\bf Di Jin}$^{\ddagger}$\quad
    {\bf Tagyoung Chung}$^{\ddagger}$\quad
    {\bf Nanyun Peng}$^{\dagger\ddagger}$
    \\
    $^{\dagger}$University of California, Los Angeles
    \quad
    $^{\ddagger}$Amazon Alexa AI\\
    {\tt \{ma, violetpeng\}@cs.ucla.edu}
    \\
    {\tt \{jiunyk, guparpit, djinamzn, tagyoung\}@amazon.com}
    \\
    {\tt shuyangg@gmail.com}
}

\begin{document}
\maketitle
\begin{abstract}
    Dialogue state tracking (DST) is an important step in dialogue management to keep track of users' beliefs. Existing works fine-tune all language model (LM) parameters to tackle the DST task, which requires significant data and computing resources for training and hosting. The cost grows exponentially in the real-world deployment where dozens of fine-tuned LM are used for different domains and tasks. To reduce parameter size and better utilize cross-task shared information, we propose to use soft prompt token embeddings to learn task properties. Without tuning LM parameters, our method drastically reduces the number of parameters needed to less than 0.5\% of prior works while achieves better low-resource DST performance.\looseness=-1
\blfootnote{$^\mathsection$Work done while at Amazon.}
\end{abstract}

\section{Introduction}
\label{sec:intro}
Dialogue state tracking (DST) that extracts structured conversation progress in a list of slot-value pairs from unstructured dialogue utterances is an essential component of a dialogue system \cite{wang-lemon-2013-simple}. 
Unlike classification-based models that pick the slot value from given candidate \cite{ye2021slot,chen2020schema}, recent works formulate DST as a conditional generation task \cite{gao-etal-2019-dialog,lin-etal-2020-mintl}, where the concatenation of dialogue history and a slot-specific prompt
are fed to generative models and the text generation output are decoded to predicted slot values \cite{ham-etal-2020-end,hosseini2020simple}. This formulation enjoys the benefit of generalizability to unseen domains and slot types beyond a defined dialogue ontology \cite{li-etal-2021-zero,peng-etal-2021-soloist}. 

General prompting methods use a textual prompt to provide task information to the LM \cite{liu2021pre,ma-etal-2023-star}. Prior works have variations that update different parameter combinations such as both LM and prompt token embeddings \cite{gao-etal-2021-making, li-liang-2021-prefix,xu-etal-2023-can-nli}, only the token embeddings of the LM \cite{zhu-etal-2021-counter-interference}, or only the prompt token embeddings \cite{lester-etal-2021-power,gu-etal-2022-ppt,vu-etal-2022-spot}. 


While there are some existing prompt-based approaches for DST with different designs of prompts such as using slot name \cite{lee2019zero,zhao-etal-2021-effective-sequence,lee-etal-2021-dialogue,su-etal-2022-multi}, slot description \cite{rastogi2020towards}, slot type \cite{lin-etal-2021-leveraging}, possible values \cite{lin-etal-2021-leveraging}, priming examples \cite{gupta-etal-2022-show} and/or slot-specific question \cite{gao-etal-2019-dialog,Zhou2019MultidomainDS,gao-etal-2020-machine,lin-etal-2021-zero,li-etal-2021-zero,wu-etal-2022-incorporating} in prompt sentences, 
they all fine-tune the entire LM along with the prompt tokens for a new domain, which requires a significant amount of training time, system resources, and annotated data \cite{clarke-etal-2022-one,sauer-etal-2022-knowledge}. 
The computing and data resource-hungry issues are more severe in the real-world deployment where LMs tuned for different domains and tasks need to be trained and hosted, and a typical dialogue system has to serve dozens of such LMs \cite{maronikolakis-schutze-2021-multidomain,strubell-etal-2019-energy,lacoste2019quantifying}. This leads to a high cost of the development and service of dialogue systems and constrains offline deployment. In addition, limited data is available for a new domain or task.\looseness=-1

\begin{figure*}[thp]
    \centering
    \includegraphics[width=\textwidth]{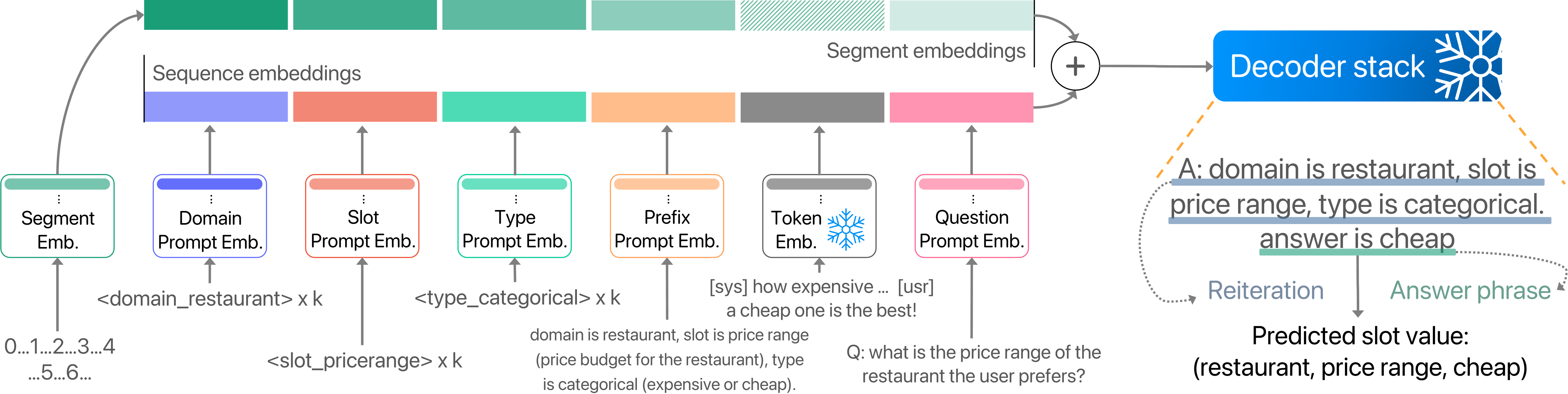}
    \caption{Model design. The snow icon indicates non-trainable parameters. Absolute positional embeddings are added together with segment embeddings and sequence embeddings, we omit it for simplicity in the illustration.}
    \label{fig:model}
\end{figure*}

We propose a \textbf{parameter-efficient} and \textbf{data-efficient} DST model for \textbf{low-resource} settings, which only needs to update 0.08\% of parameters compared with the previous best model, by keeping LM parameters frozen and introducing soft prompt tokens to represent task properties of different slots.  \figref{fig:model} gives an overview of our model. 
The only prior work we are aware of that only updates prompt token embeddings and thus parameter-efficient is \citet{zhu-etal-2022-continual}, but it focuses on continual domain adaptation and with a significant amount of training data. 
\looseness=-1

Our design introduces three techniques that are generalizable to other generative-based information extraction models. 
1) \textbf{Task-specific parameters}:
\textit{task prompt tokens} are introduced to specifically learn domain, slot and slot type information so that the model behaves according to the task; \textit{word-mapping prompt tokens} enable us to obtain task knowledge contained in natural language instruction and optimize human-created prompts with continuous embedding space. 
2) \textbf{Task metadata in objective}:
we introduce the reiteration technique in the target sequence in order to include explicit task signals in the text generation objective. 
3) \textbf{Distinguishing segments}:
segment embeddings help the model identify the prompt segment, dialogue speakers, and question partition.
Our proposed method enables much more efficient dialogue system deployment as only one LM needs to be hosted and inference for different domains could be realized by feeding domain-specific prompt token embeddings into the transformer stack. 

Experiments on MultiWOZ 2.0 show that our method achieves better performance on low-resource DST with orders of magnitude fewer parameters. We further conduct ablation studies, error analysis, and examine the semantic information shown in the prompt tokens. We observe that our model is more specialized in predicting categorical slot values, is more conservative for slots with free output space and introduces more hallucination errors for categorical slots.
\looseness=-1

\section{Method}
\label{sec:method}
We introduce task definition (\secref{sec:task_definition}), overall framework (\secref{sec:gen_framework}) and soft prompt designs (\secref{sec:soft_prompt}) in this section.

\subsection{Task Definition}
\label{sec:task_definition}
The goal is to construct a belief state with $|S|$ pairs of slot and value at a certain turn in a multi-turn conversation. All the turns up to the query turn are dialogue history, and slot-specific information (\ie name, description, value candidates, question and type of the slot) is provided.\footnote{We show slot-specific info in \appref{app:slot_type_definitions} and \appref{app:question_prompt}.
Value candidates are from dialogue ontology.
}
\subsection{Generative Seq2seq Framework}
\label{sec:gen_framework}
We use a decoder-only pre-trained language model (PLM) GPT-2 \cite{radford2019language} as the backbone to provide language and commonsense knowledge, rather than an encoder-decoder model because of its superior performance \cite{li-etal-2021-zero}. 
To get a belief state at a certain turn, we create $|S|$ data instances to predict the slot value for each slot.
\figref{fig:model} demonstrates the design and a sample query.

\mypar{Input sequence.} We construct the input sequence by concatenating the following segments: 
1) \textit{Task prompt tokens for domain, slot and type}, each has $k$ prompt tokens and they are shared among instances with the same domain, slot or type; 
2) \textit{Prefix}, a short sentence containing slot description, names of domain, slot, and type, and all possible candidates if the query slot is categorical;
3) \textit{Dialogue history}, in which \texttt{[sys]} and \texttt{[usr]} tokens are used to indicate the speaker; and
4) \textit{Question}, human-written question about the slot.

\mypar{Target sequence and reiteration.} We introduce the reiteration technique in the target sequence as shown in \figref{fig:model} and generate task information before the answer phrase. We include the verbalized slot information as a ``domain is \texttt{domain name}, slot is \texttt{slot name}, type is \texttt{type name}'' phrase in the expected output sequence. 
By doing so, we require the model to optimize to remember the task information explicitly before generating the answer phrase, while using a consistent text generation cross-entropy loss. 
This technique allows the model to optimize upon both the answer and the sentence containing slot metadata, and explicitly learn the task information.

\begin{table*}[ht!]
\begin{center}
\resizebox{\linewidth}{!}{
{
\small
\setlength\tabcolsep{2pt}
\begin{tabular}{lr|
cccccc|
cccccc|
cccccc
}
\toprule
\multirow{2}{*}[-4pt]{Model} 
& \multirow{2}{*}[-4pt]{Params\#}
& 5 & 10 & 20 & 1\% & 5\% & 10\% & 5 & 10 & 20 & 1\% & 5\% & 10\% & 5 & 10 & 20 & 1\% & 5\% & 10\% \\  \cmidrule{3-20}
& 
& \multicolumn{6}{c|}{Attraction (3 slots, 1\% = 27 conv.)} 
& \multicolumn{6}{c|}{Hotel (10 slots, 1\% = 33 conv.)}
& \multicolumn{6}{c}{Restaurant (7 slots, 1\% = 38 conv.)}
\\
\midrule
TRADE &
&---&---&---&---&52.19&58.46
&---&---&---&---&31.93&41.29
&---&---&---&---&47.31&53.65
\\
DSTQA &
&---&---&---&---&51.58&61.77
&---&---&---&---&33.08&\textbf{49.69}
&---&---&---&---&35.33&54.27
\\
T5DST & 60M
&4.77&21.93&30.57&40.68&52.12&60.13
&8.19&13.46&17.94&18.63&38.76&46.13
&13.80&19.51&22.79&29.47&\textbf{53.32}&58.44
\\
\citeauthor{lee-etal-2021-dialogue} & 60M
& 6.33 & 19.12 & 34.53 & 37.56 & 54.34 & 58.75
& 9.31 & 15.76 & 22.07 & 24.41 & \textbf{40.11} & 42.98
& 15.87 & 19.66 & 22.15 & 30.96 & 48.94 & \textbf{58.59}
\\
\citeauthor{li-etal-2021-zero} & 335M
&7.90&27.09&35.63&42.18&49.13&60.85 
&12.49&15.15&19.44&24.04&37.88&46.47
& 17.27 & 22.30 & 25.68&30.70&49.75&58.50
\\
\midrule
Ours & 271K
& \textbf{33.56} & \textbf{39.41} & \textbf{45.75} & \textbf{47.28} & \textbf{56.99} & \textbf{63.61}
& \textbf{15.63} & \textbf{18.18} & \textbf{22.50} & \textbf{33.01} & 38.24 & 45.60
& \textbf{19.76} & \textbf{25.72} & \textbf{27.65} & \textbf{34.40} & 50.81 & 55.79
\\
\toprule 
\multirow{2}{*}{} 
& \multirow{2}{*}{}
& \multicolumn{6}{c|}{Taxi (4 slots, 1\% = 15 conv.)}
& \multicolumn{6}{c|}{Train (6 slots, 1\% = 29 conv.)}
& \multicolumn{6}{c}{Average}
\\
\midrule
TRADE &
&---&---&---&---&59.03&60.51
&---&---&---&---&48.82&59.65
&---&---&---&---&47.86&54.71
\\
DSTQA &
&---&---&---&---&58.25&59.35
&---&---&---&---&50.36&61.28
&---&---&---&---&45.72&57.27
\\
T5DST & 60M
&48.22&53.74&58.27&58.19&59.23&69.03
&12.31&21.93&36.45&43.93&69.27&69.48
&17.46&26.11&33.20&38.18&54.54&60.64
\\
\citeauthor{lee-etal-2021-dialogue} & 60M
& 45.32 & 49.93 & 58.58 & 58.52 & 60.77 & \textbf{71.23}
& 13.57 & 25.02 & 38.52 & 50.26 & 69.32 & 69.72
& 18.08 & 25.90 & 35.17 & 40.34 & 54.70 & 60.25
\\
\citeauthor{li-etal-2021-zero} & 335M
&50.99&57.47&58.49&58.26&\textbf{61.68}&69.23
&17.56&27.42&39.27&45.32&\textbf{71.69}&73.45
&21.24&29.89&35.70&40.10&54.03&\textbf{61.70}
\\
\midrule
Ours & 271K
& \textbf{51.11} & \textbf{59.63} & \textbf{60.89} & \textbf{60.33} & 61.63 & 63.00
& \textbf{18.95} & \textbf{30.95} & \textbf{50.34} & \textbf{52.05} & 69.51 & \textbf{75.00}
& \textbf{27.80} & \textbf{34.78} & \textbf{41.43} & \textbf{45.41} & \textbf{55.44} & 60.60

\\
\bottomrule
\end{tabular}
}
}
\caption{
Overall performance.
Detailed parameter counts are in \appref{app:detailed_parameter_count},
variances are in \appref{fewshot_variance}.
}
\label{table:overall}
\end{center}
\end{table*}

\mypar{Segment embeddings.} The input sequence contains segments with diverse formats and they are quite different from the format used in the pre-training phase of the LM. We divide the input sequence into segments, including five prompt segments, the system turns, the user turns and the answer segment. Tokens within a segment are assigned the same segment ID. Segment embeddings, which have the same length as the input sequence, are added with sequence embeddings and positional embeddings. 
We randomly initialize the embeddings of segment IDs and update them during training.

\mypar{Training and inference.} 
We pass the combined embeddings to the decoder stack to calculate the likelihood over the vocabulary. We use the cross-entropy loss with a regularization term to constrain the scale of prompt token embeddings following
    $L = CE + \lambda\|PE^\prime - PE\|_{2}^{2}$
where $\lambda$ is a weighting factor, and $PE^\prime$ and $PE$ are updated and initialized prompt token embeddings \cite{muller-etal-2022-shot}.
Parameters of the PLM are frozen, and
only prompt and segment embeddings are updated 
with Adam optimizer. During inference, we generate the output autoregressively with greedy decoding, and extract the answer with a rule-based function.

\subsection{Soft Prompt Tokens} 
\label{sec:soft_prompt}

\mypar{Prompt segments.}
We use two kinds of prompt tokens. 
\textit{Task prompt tokens} are chosen according to the task's metadata, and used in the domain, slot and type prompt segments. \textit{Word-mapping prompt tokens} are mapped from existing tokens in the prefix and question parts and used to replace normal tokens.
In other words, task and word-mapping prompt tokens are shared across instances with the same task and instances using the same words respectively. We concatenate embeddings of each prompt segment (obtained by separate embedding matrices) with dialogue history embeddings (obtained by the frozen token embedding matrix) to form sequence embeddings.

\mypar{Prompt initialization.}
To boost the performance in the low-resource setting, we use the pre-trained token embeddings to initialize the soft prompt token embeddings. The token embeddings from PLM are used to represent word semantics for language understanding, while the soft prompt tokens are used to represent task information initialized by task-related semantic meanings. We initialize a task prompt token by embedding of a randomly chosen token from its domain, slot or slot type name. Word-mapping prompt tokens are initialized with the embedding of the mapped word.

\section{Experimental Setup}
\mypar{Dataset.} We experiment on dialogues of five domains (\ie attraction, hotel, restaurant, train, taxi) in MultiWOZ 2.0 \cite{budzianowski-etal-2018-multiwoz}.
\looseness=-1

\mypar{Settings.}
We evaluate using the low-resource few-shot DST task. We take 5, 10, 20, 1\%, 5\% and 10\% of training conversations to train, and evaluate on the full test set of each target domain.\footnote{\appref{app:experiment_setting_details} and \apprefsimple{app:baseline_models_details} show experimental setting details.}

\mypar{Evaluation metrics.} 
Joint Goal Accuracy (JGA) represents the proportion of \textit{turns} with \textit{all} slots predicted correctly, and Slot Accuracy (SA) reflects the proportion of correct \textit{slots}.
If a slot is empty at a certain turn (for example, no related information is mentioned), the model needs to predict ``none''.
A slot value is only correct if it matches exactly with the ground-truth value.

\mypar{Baseline models.}
We compare with the following works.\footnote{We are not comparing with prompt-based DST works that jointly train with other tasks for a fair comparison.} 1) TRADE \cite{wu-etal-2019-transferable}: GRU-based model with copy mechanism; 2) DSTQA \cite{Zhou2019MultidomainDS}: QA-style model using ELMo representation; 3) T5DST \cite{lin-etal-2021-leveraging}: T5-based generative model with slot type as prompt; 4) \citet{lee-etal-2021-dialogue}: T5-based generative model with slot description and possible slot values as prompt; 5) \citet{li-etal-2021-zero}: GPT-2 based QA-style generative model with manually created questions. The entire language model is updated for T5DST, \citeauthor{lee-etal-2021-dialogue} and \citeauthor{li-etal-2021-zero}, and they represent the performance of prompt-based DST works. \appref{app:result_frozen_baseline} shows comparison with baselines' frozon LM variation.\looseness=-1
\section{Experimental Results}
\label{sec:eval}
\mypar{Overall results.} We show the overall few-shot experimental results in \tbref{table:overall}. 
Although our model uses only 0.08\% and 0.45\% of parameters compared with baselines, it still achieves higher JGA than all baseline models when using 1\% or less training data across all domains. Especially we observe around 5, and 9 points JGA increases for the \texttt{attraction} and \texttt{hotel} domains compared with existing best models with 1\% training data. 
In the \texttt{attraction} domain with 3 unique slots, our model trained using 5 dialogues performs on par with the previous best model using 20 dialogues. Our model shows its superiority especially when the amount of unique tasks is small.
Using 5\% and 10\% data, our model performs comparably with existing best models with small gaps.
\begin{figure}[h]
    \centering
    \includegraphics[width=\columnwidth]{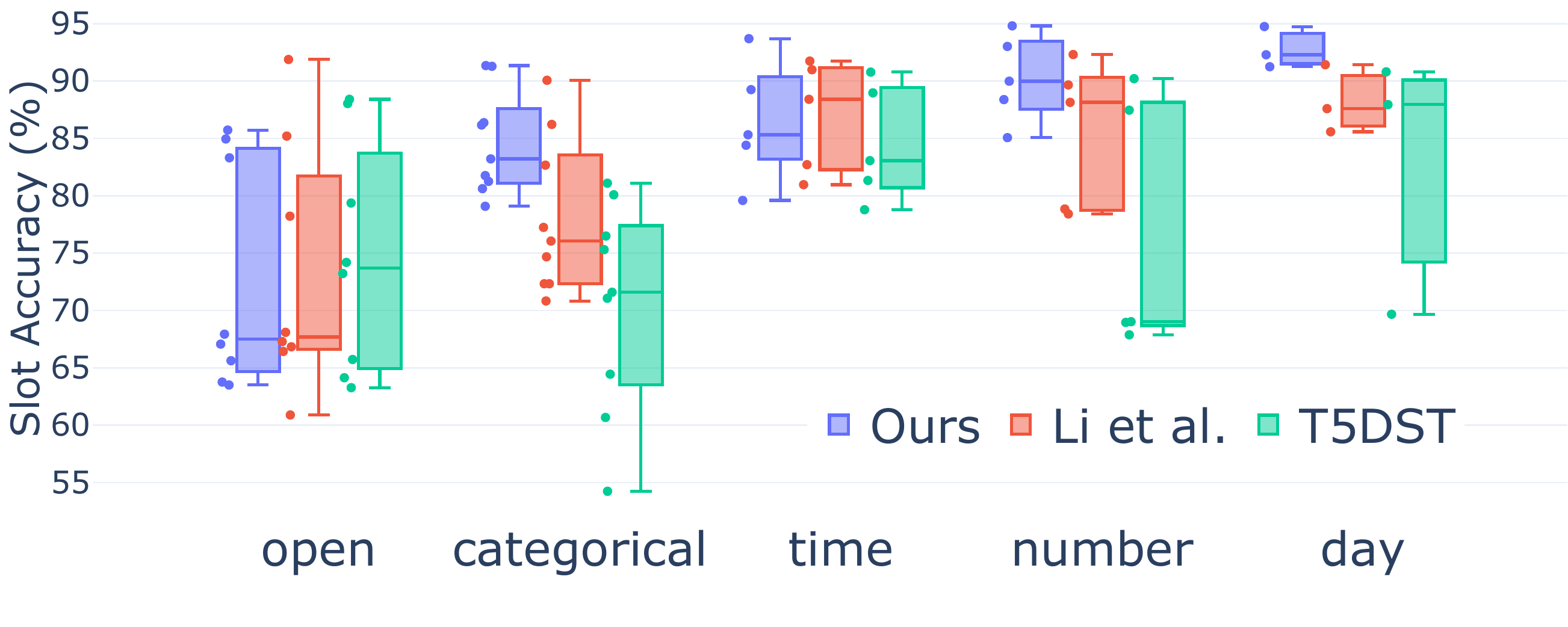}
    \caption{Slot accuracies across slot types using 1\% training data, each dot represents a unique slot.}
    \label{fig:slot_type}
\end{figure}

We demonstrate the performance of slots with different types in \figref{fig:slot_type} across all five domains in \figref{fig:slot_type} compared with two generative baselines \citeauthor{li-etal-2021-zero} and T5DST \cite{lin-etal-2021-leveraging}. These slot types are defined in \tbref{table:slot_types}.
We observe the worst performance in \textsc{open} slots, which could be explained by the larger output candidate space.\footnote{A SA vs ontology size analysis is in \appref{app:onto_size}.}
Breaking down slot type to more fine-grained type lead to a better result (considering \textsc{day} as a separate type rather than \textsc{categorical} type, \textsc{number} and \textsc{time} as separate types rather than \textsc{open} type).
Compared with baselines, our model performs comparably on \textsc{open} and \textsc{time} slots, but is more superior for \textsc{categorical}, \textsc{number} and \textsc{day} slots.\footnote{SA for each slot and comparisons are in \appref{app:performance_by_slots}.}

\mypar{Ablation study.}
In \tbref{table:ablation},
removing the slot segment (Line 2) leads to the largest performance drop among the three task prompt segments (L1-3), as slot is the most fine-grained task categorization. Prefix (L5) is more important than the question prompt (L4), which contains more metadata and parameters.
The model without segment embedding (L6) has on average 7.8 points JGA drop, indicating the effectiveness of the segment embedding. We also observe an almost 2 points JGA drop (and an even larger drop with fewer training dialogues shown in \appref{app:ablation_reiteration}) without reiteration (L7), which shows the helpfulness of including explicit task information in the learning objective.
Note that even without reiteration, our model performs better than all baselines using 1\% training data.
\begin{table}[th]
\begin{center}
\resizebox{\linewidth}{!}{
{
\small
\setlength\tabcolsep{2pt}
\begin{tabular}{clH
cHH
cHH
cHH
cHH
cHH
c}
\toprule
\multirow{1}{*}{\#} 
& \multirow{1}{*}{Model} 
& \multirow{1}{*}{\# Params}
& \multicolumn{3}{c}{Attr.} 
& \multicolumn{3}{c}{Hotel}
& \multicolumn{3}{c}{Rest.}
& \multicolumn{3}{c}{Taxi}
& \multicolumn{3}{c}{Train}
& Avg
\\
\midrule
1 & w/o domain &
& 44.22 & &
& 28.16 & &
& 29.78 & & 
& 60.27 & &
& 50.01 & &
& 42.49
\\
2 & w/o slot &
& 46.64 & &
& 26.55 & &
& 24.35 & &
& 51.11 & &
& 45.11 & &
& 38.75
\\
3 & w/o type &
& 45.30 & &
& 25.26 & &
& 33.65 & &
& 59.89 & &
& 51.91 & &
& 43.20
\\
4 & w/o question &
& 45.08 & &
& 32.26 & &
& 33.30 & &
& 59.63 & &
& 51.60 & &
& 44.37
\\
5 & w/o prefix &
& 42.98 & &
& 28.78 & &
& 31.54 & &
& 57.72 & &
& 47.00 & &
& 41.60
\\\midrule
6 & w/o segment emb. &
& 34.35 & &
& 23.18 & &
& 27.33 & &
& 59.69 & &
& 43.30 & &
& 37.57
\\
\midrule
7 & w/o reiteration &
& 45.08 & &
& 27.57 & &
& 33.48 & &
& 59.89 & &
& 51.08 & &
& 43.42 
\\
\midrule
8 & Full model & 390K 
& 47.28 & &
& 33.01 & &
& 34.40 & &
& 60.33 & &
& 52.05 & &
& 45.41
\\
\bottomrule
\end{tabular}
}
}
\caption{
Ablation study using 1\% training data (JGA).
}
\label{table:ablation}
\end{center}
\end{table}

\begin{figure}[htp]
    \centering
    \includegraphics[width=\columnwidth]{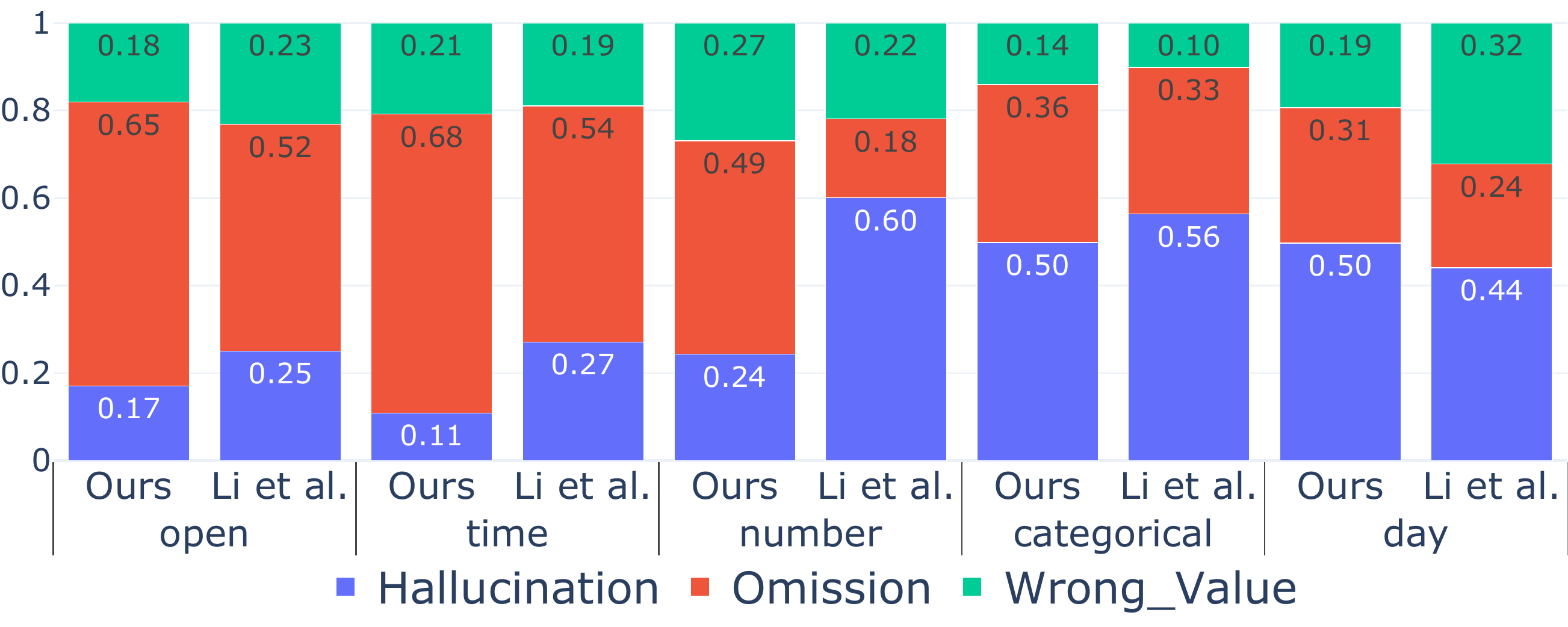}
    \caption{Error distribution across slot types}
    \label{fig:error_by_slot_type}
\end{figure}

\mypar{Error and qualitative analysis.}
We categorize error cases as: 1) hallucination: predicting value for an empty slot; 2) omission: predicting ``none'' for a non-empty slot; 3) wrong value: predicting wrong real value for a non-empty slot \cite{gao-etal-2020-machine}. 
\figref{fig:error_by_slot_type} shows\mdma{ the error distribution in terms of the proportion of each error category.} The general \textsc{open} slots (including \textsc{time} and \textsc{number}) have \mdma{relatively} more omission errors, while the general \textsc{categorical} slots have \mdma{relatively} more hallucination errors. Our model is more conservative for \textsc{open} slots compared with \citeauthor{li-etal-2021-zero}.\footnote{\mdma{Our model produces \textit{relatively} larger proportion of omission error than \citeauthor{li-etal-2021-zero}, but it generates a reasonable amount of not-none values for non-empty slots as explained in \appref{app:more_error_analysis}.}}

We then investigate semantic information contained in the learned prompt tokens by selecting the most changed prompt tokens and producing the closest tokens with the smallest cosine similarity between the learned prompt token embedding and frozen token embeddings of the PLM. We show the result for the \texttt{attraction} domain in \tbref{table:cloest_tokens}, and for all domains in \appref{app:cloest_tokens}. The closest tokens are mostly variations or semantically similar tokens of the expected meanings of prompt tokens.

\begin{table}[ht]
\begin{center}
\resizebox{\linewidth}{!}{
{
\small
\setlength\tabcolsep{2pt}
\begin{tabular}{HHll}
\toprule
Domain & segment
& Prompt token
& Closest tokens
\\
\midrule
\multirow{10}{*}{Attraction} & \multirow{2}{*}{Domain}
& <domain\_attraction\_4> & raction; ractions; racted
\\
& \multirow{2}{*}{Slot}
& <slot\_name\_2>& name; Name; names
\\
& \multirow{2}{*}{Type}
& <type\_open\_3> & open; Open; opened
\\
& \multirow{2}{*}{Prefix}
& special & special; Special; statistical
\\
& \multirow{2}{*}{Question}
& Q & answer; Answer; answered
\\
\bottomrule
\end{tabular}
}
}
\caption{
Closest tokens for the most changed prompt tokens in five prompt segments for the attraction domain.
}
\label{table:cloest_tokens}
\end{center}
\end{table}




\section{Related Works}
\label{sec:relatedworks}
Two lines of work are closely related to our work.
\mypar{Dialogue State Tracking.}
DST aims to extract slots and values to capture user intents from dialogues.
In terms of model design, \textit{classification-based models} pick the slot value from candidates \cite{ye2021slot,
gao-etal-2020-machine,ma-etal-2019-implicit,
chen2020schema}, which assumes that the dialogue ontology is pre-defined and cannot be generalized \cite{chen2020schema}. 
\textit{Generation-based models} directly generate slot values to handle unseen domains and values \cite{ham-etal-2020-end,hosseini2020simple,
gao-etal-2019-dialog,gao-etal-2020-machine,
lin-etal-2020-mintl,
peng-etal-2021-soloist}.
In terms of knowledge sources, 
some methods create synthesized dialogues with human heuristics to do data augmentation for the target domain \cite{campagna-etal-2020-zero, hou2020c2c} which require expensive human costs. 
Recent works transfer knowledge from data-rich domains with slot description \cite{lee2019zero, rastogi2020towards}, slot type, possible values \cite{lin-etal-2021-leveraging}, 
possible values \cite{lee-etal-2021-dialogue}, 
task  constraint \cite{mi2021cins}, similarity functions between slots and values 
\cite{wu-etal-2020-tod}, and meta-learning \cite{dingliwal-etal-2021-shot}, while the availability of related source domains constrains their generalizability. Some works transfer from other tasks like Reading Comprehension \cite{gao-etal-2019-dialog, gao-etal-2020-machine, lin-etal-2021-zero}.  
We take inspiration and use a transformer-based generative model with slot metadata but using much less trainable parameters.
\looseness=-1

\mypar{Prompting methods.}
Recent works use a textual prompt 
as a part of the input sequence 
to provide task information to the LM \cite{liu2021pre}. 
The prompt can be chosen by experts \cite{radford2019language,lu-etal-2023-multihop,lu-etal-2022-summarization}, learned as discrete readable tokens \cite{shin2020eliciting} or continuous embeddings \cite{qin-eisner-2021-learning}.
The textual prompt can also contain a few examples known as ``in-context learning'' without tuning LM \cite{brown2020language}. 
Some works fine-tune both LM and 
prompt parameters \cite{gao-etal-2021-making, qin-eisner-2021-learning, li-liang-2021-prefix,ma-etal-2022-dice,hsu-etal-2022-degree} or only token embeddings \cite{zhu-etal-2021-counter-interference}.
Works like \citet{lester-etal-2021-power,gu-etal-2022-ppt,vu-etal-2022-spot} show that freezing PLM 
and only tuning learnable soft prompts, known as ``prompt tuning'', is competitive with fine-tuning while using much less parameters. 
For the DST task, \citet{lee-etal-2021-dialogue} use slot information as prompt and fine-tune PLM, 
\citet{zhu-etal-2022-continual} use prompt tuning for continual domain adaptation,
both requiring a significant amount of training data. 
For the low-resource DST task, \citet{yang2022prompt} introduce two-way prompts but need to fine-tune LM, which is not parameter-efficient.\looseness=-1

\section{Conclusion and Future Work}
\label{sec:conclusion}
We propose a parameter-efficient DST model using prompt tuning, and it represents tasks with soft prompt tokens with segment awareness and reiteration. 
Our model achieves state-of-the-art low-resource DST performance with less than 0.5\% parameters compared with fine-tuning LM.
We plan to further investigate 
prompt aggregation.

\section*{Limitations}

There are several limitations to our work. Firstly, the proposed model is more sensitive to hyper-parameters such as the number of prompt tokens and learning rate than existing methods that fine-tune LM. Therefore, it would require additional parameter searching efforts to obtain the best performance. Secondly, our model is designed for and evaluated in English-only conversations, and applying our technique to other languages or code-switching scenarios might lead to performance decay. Finally, our experimental result shows that our proposed prompt tuning method works better than fine-tuning LM when there are fewer unique tasks to be optimized. Therefore, our method might not work well on a more diverse dataset.

\section*{Ethics Statement}
We do not see an immediate negative impact of the proposed method and do not see biased predictions made by our model. Our method is based on a pre-trained generative language model and trained on an open DST dataset, thus bias contained in the corpus for pre-training and the DST dataset might propagate to prediction outputs of our model. 
Human validation of the prediction results and their fairness needs to be conducted before our proposed model is used in production and other real-world applications. Our proposed model does not increase energy and carbon costs but will potentially reduce them due to its data and parameter efficiency.

\section*{Acknowledgments}

Many thanks to Sidi Lu, Tanmay Parekh, and Sarik Ghazarian for internal reviews, to members at Amazon Alexa AI, PLUS lab and UCLA-NLP for suggestions, and to the anonymous reviewers for their feedback.

\bibliography{anthology,ma}
\bibliographystyle{acl_natbib}

\clearpage

\appendix

\section{Design Details}
\label{appendix:appendix-1}

\subsection{Slot Type Definitions}
\label{app:slot_type_definitions}
The slot types are defined according to output space and the number of possible answers. The slot types are defined in \tbref{table:slot_types}.



\begin{table}[h]
\setlength\tabcolsep{2pt}
    \centering
    {\small
    \begin{tabularx}{\columnwidth}{lX}
    \toprule
        Slot Types & Slots \\ \midrule
        Categorical & attraction-area, hotel-area, hotel-internet, hotel-parking, hotel-price range, hotel-type, restaurant-area, restaurant-price range, train-day \\\midrule
        Day & hotel-book day, restaurant-book day \\ \midrule
        Number & hotel-book people, hotel-book stay, hotel-stars, restaurant-book people, train-book people \\\midrule
        Open & attraction-open, attraction-type, hotel-name, restaurant-food, taxi-departure, taxi-destination, train-departure, train-destination \\\midrule
        Time & restaurant-book time, taxi-arrive by, taxi-leave at, train-leave at
        \\\bottomrule
    \end{tabularx}
    }
    \caption{Slot type definitions.}
    \label{table:slot_types}
\end{table}

\subsection{Question Prompt and Description}
\label{app:question_prompt}
We show questions (used as question prompts) and description (as part of prefix prompt) for each slot in \tbref{table:app_question}. Slot descriptions are from MultiWOZ 2.2 dataset \cite{zang-etal-2020-multiwoz}.

\subsection{Detailed Parameter Count}
\label{app:detailed_parameter_count}
The average parameter count across all domains is 271K.
We show detailed parameter count for each domain in \tbref{table:param_num_count}. The parameters needed for each domain vary because the question and prefix prompt can map to a different set of prompt tokens for each domain. The parameters needed for each domain are calculated by adding prompt token embedding size ($prompt\ token\ count \times 1024$) with segment embedding size ($8 \times 1024$ given 8 segments).
\begin{table}[h]
\begin{center}
{
\small
\setlength\tabcolsep{2pt}
\begin{tabular}{l
r
r
r
r
r
r}
\toprule
 
& Attr.
& Hotel
& Rest.
& Taxi
& Train
\\

\midrule
Domain
& 5 & 20 & 20 & 10 & 10
\\
Slot
& 15 & 200 & 140 & 40 & 60
\\
Type
& 10 & 80 & 100 & 20 & 40
\\
Question
& 20 & 46 & 36 & 19 & 27
\\
Prefix
& 60 & 117 & 84 & 29 & 76
\\
\midrule
All
& 110 & 463 & 380 & 118 & 213
\\
Params \#
& 120832 & 482304 & 397312 & 129024 & 226304
\\
\bottomrule
\end{tabular}
}

\caption{
The number of parameters needed for each domain. We list the number of prompt tokens needed for each prompt segment, all prompt tokens needed and the ultimate parameter count.
}
\label{table:param_num_count}
\end{center}
\end{table}
\section{Additional Experimental Results}

\subsection{Additional Ablation Study for Reiteration}
\label{app:ablation_reiteration}
We show an additional ablation study to investigate the effect of introducing the reiteration technique in \tbref{table:ablation_reiteration}. 
We observe that the reiteration technique can lead to a significant increase in performance, especially with fewer amount of training dialogues. When there are limited training data, reiteration can help the model learn task boundaries among each slot faster and better.
\begin{table}[ht]
\begin{center}
{
\small
\setlength\tabcolsep{2pt}
\begin{tabular}{clH
rHH
rHH
rHH
rHH
rHH
c}
\toprule
\multirow{1}{*}{Few-shot} 
& \multirow{1}{*}{Model} 
& \multirow{1}{*}{\# Params}
& \multicolumn{3}{c}{Attr.} 
& \multicolumn{3}{c}{Hotel}
& \multicolumn{3}{c}{Rest.}
& \multicolumn{3}{c}{Taxi}
& \multicolumn{3}{c}{Train}
& Avg
\\ \midrule
\multirow{2}{*}{5}  & w/o reit. &
& 22.16 & &
& 12.09 & &
& 16.67 & &
& 47.68 & &
& 4.97 & &
& 20.71
\\
 & w/ reit. & 390K 
& 33.56 & &
& 15.63 & &
& 19.76 & &
& 51.11 & &
& 18.95 & &
& 27.80
\\
\midrule
\multirow{2}{*}{10}  & w/o reit. &
& 23.08 & &
& 12.39 & &
& 13.75 & &
& 56.39 & &
& 9.26 & &
& 22.97
\\
 & w/ reit. & 390K 
& 39.41 & &
& 18.18 & &
& 24.72 & &
& 59.63 & &
& 30.95 & &
& 34.58 
\\
\midrule
\multirow{2}{*}{1\%}  & w/o reit. &
& 45.08 & &
& 27.57 & &
& 33.48 & &
& 59.89 & &
& 51.08 & &
& 43.42 
\\
 & w/ reit. & 390K 
& 47.28 & &
& 33.01 & &
& 34.40 & &
& 60.33 & &
& 52.05 & &
& 45.41
\\
\bottomrule
\end{tabular}
}

\caption{
Ablation study for the reiteration technique.
}
\label{table:ablation_reiteration}
\end{center}
\end{table}

\subsection{Performance vs Ontology Size}
\label{app:onto_size}

\begin{figure}[htp]
    \centering
    \includegraphics[width=\columnwidth]{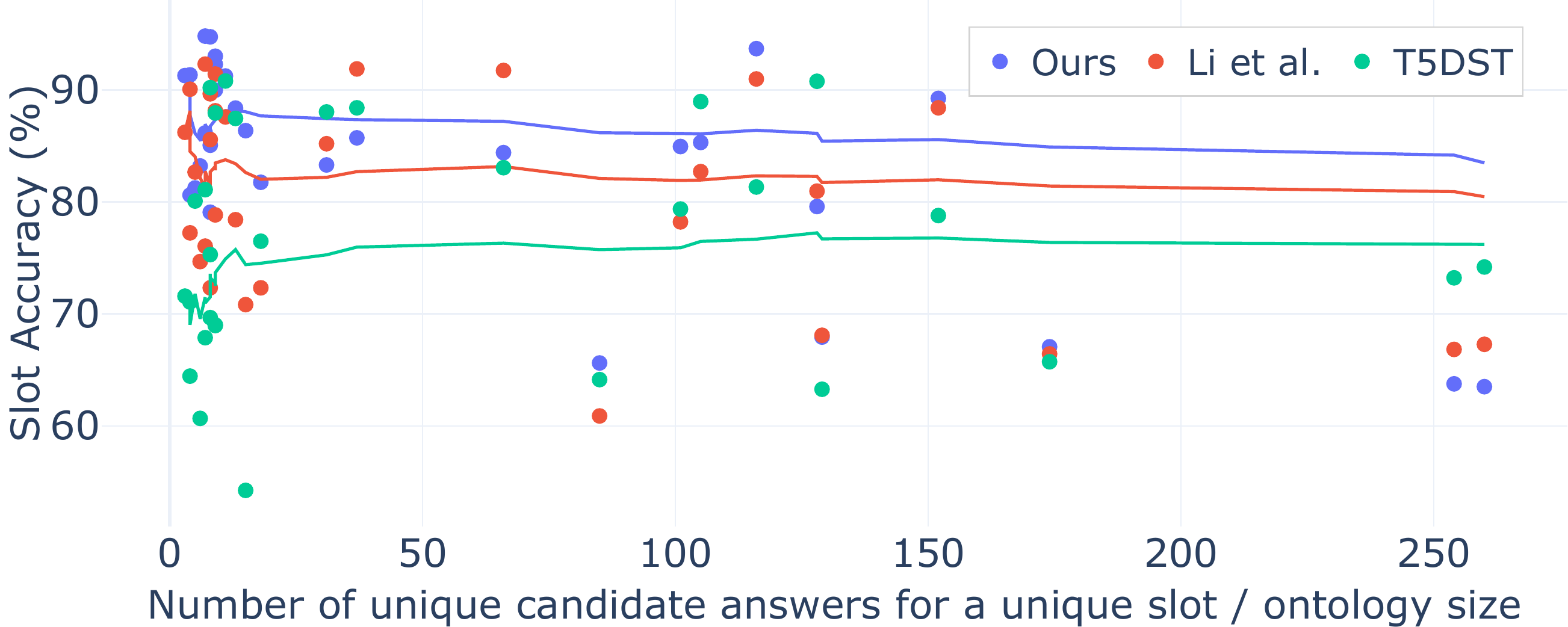}
    \caption{Performance for slots with different ontology sizes}
    \label{fig:onto_size}
\end{figure}

We investigate the relationship between performance and the number of unique candidate answers (ontology size) using 1\% target domain training data and \figref{fig:onto_size} demonstrates the result with trendlines created by expanding average algorithm for each model. We also show the performance of two generative baseline models for comparison. We observe that the performance of all three models drops when the ontology size grows. For most ontology sizes, our model outperforms \citeauthor{li-etal-2021-zero} and T5DST \cite{lin-etal-2021-leveraging}.

\subsection{Performance by Slots}
\label{app:performance_by_slots}

\begin{figure*}[thp]
    \centering
    \includegraphics[width=\textwidth]{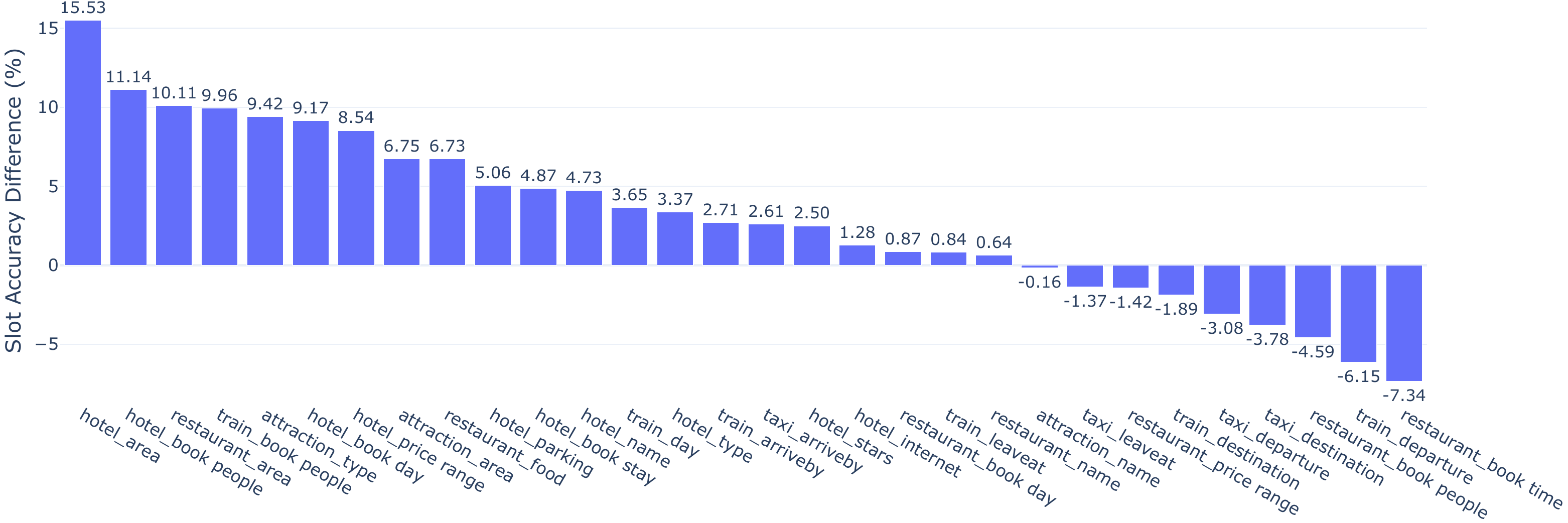}
    \caption{Slot accuracy difference between our model and \cite{li-etal-2021-zero}, a positive value indicates our model is better.}
    \label{fig:SA_diff}
\end{figure*}

To better understand the pros and cons of the prompt tuning method compared with fine-tuning LM, we show the slot accuracy difference for all unique slots training with 1\% target domain data in \figref{fig:SA_diff}. Our model outperforms \citet{li-etal-2021-zero} the most in \textsc{categorical}-type ``area'' slots, and \textsc{number}-type slots ``book people'', all with at least 10\% higher accuracy. Our model falls behind in 9 out of 30 slots, especially for the ``restaurant: book time'' slots.

\begin{figure}[htp]
    \centering
    \includegraphics[width=\columnwidth]{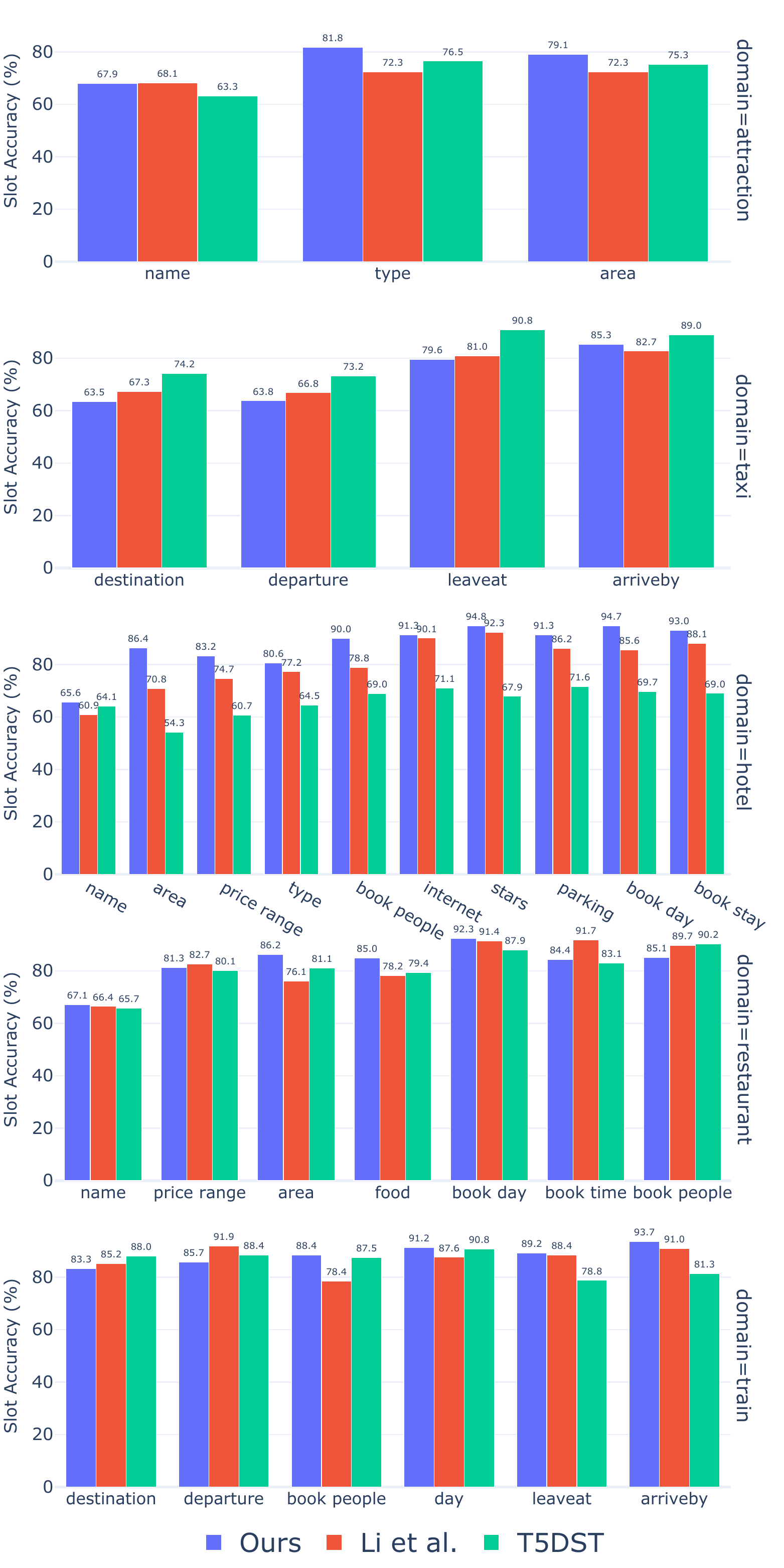}
    \caption{Slot accuracy for each slot across different domains}
    \label{fig:SA_slots}
\end{figure}
We show the slot accuracy for each unique slot in all five domains using 1\% target domain training data with comparisons to generative baselines in \figref{fig:SA_slots}.

\subsection{Closest Tokens of Learned Prompt Tokens}
\label{app:cloest_tokens}
\tbref{table:app_cloest_tokens} shows the full list of closest tokens for the most updated prompt tokens of each prompt segment in all five domains. We produce the closest tokens with the smallest cosine similarity between the learned prompt token embedding and frozen token embeddings of the PLM. 


\subsection{Variances of the Few-Shot Experimental Results}
\label{fewshot_variance}

The variances of the experimental results reported in \tbref{table:overall} (in the order of using 5, 10, 20 conversations and 1\%, 5\%, 10\% of training data): 
\begin{itemize}
    \item Attraction: 0.27, 0.33, 0.35, 0.30, 0.22, 0.32
    \item Hotel: 0.52, 0.49, 0.55, 0.57, 0.61, 0.58
    \item Restaurant: 0.63, 0.72, 0.81, 0.79, 0.83, 0.80
    \item Taxi: 0.54, 0.61, 0.54, 0.48, 0.52, 0.69
    \item Train: 0.68, 0.72, 0.73, 0.52, 0.49, 0.55
\end{itemize}

\mdma{
\subsection{Comparison with Frozen LM Version of Baselines}
\label{app:result_frozen_baseline}

Since there are many design choices to make to create a frozen LM version of the baselines (such as whether to add prefix prompt tokens, how to map tokens in the prompt segment to the underlying parameters etc), such variations would almost become new models. In our experiments, we show that our model outperforms existing models (optimizing all parameters) in low-resource settings, and we are confident that our model outperforms their frozen LM version with even larger gaps given the assumption that simply removing trainable parameters hurts the performance. 
We quantify such gaps by comparing the frozen and unfrozen version of the baseline \citeauthor{li-etal-2021-zero} with our model in \tbref{table:result_frozen_baseline}.
\begin{table}[h]
\begin{center}
{
\small
\setlength\tabcolsep{2pt}
\begin{tabular}{lrrrrr}
\toprule
Model
& Attr.
& Hotel
& Rest.
& Taxi
& Train
\\
\midrule
\citeauthor{li-etal-2021-zero} (frozen LM)
& 29.16
& 14.81
& 15.14
& 47.56
& 35.77
\\
\citeauthor{li-etal-2021-zero}
& 42.18
& 24.04
& 30.70
& 58.26
& 45.32
\\
Ours
& 47.28
& 33.01
& 34.40
& 60.33
& 52.05
\\
\bottomrule
\end{tabular}
}

\caption{
Comparison with the frozen LM variation of the baseline. JGA (\%) using 1\% training data for each domain.
}
\label{table:result_frozen_baseline}
\end{center}
\end{table}
}

\mdma{
\subsection{More about Error Analysis}
\label{app:more_error_analysis}

\figref{fig:error_by_slot_type} shows the error distribution in terms of the proportion of each error category, rather than the absolute error case counts. Though in \figref{fig:error_by_slot_type}, the omission error produced by our model takes the larger proportions in all five slot types compared with \citeauthor{li-etal-2021-zero}, our model actually makes fewer absolute omission errors than \citeauthor{li-etal-2021-zero} in the ``categorical'' and ``day'' slot types, as shown in \tbref{table:absolute_error_counts}.

\begin{table}[h]
\begin{center}
{
\small
\setlength\tabcolsep{4pt}
\begin{tabular}{lrrrrr}
\toprule
Model
& Open
& Time
& Number
& Categorical
& Day
\\
\midrule
\citeauthor{li-etal-2021-zero}
& \textbf{12.9}
& \textbf{6.2}
& \textbf{2.5}
& 7.2
& 2.8
\\
Ours
& 16.4
& 8.3
& 4.8
& \textbf{5.6}
& \textbf{2.3}
\\
\bottomrule
\end{tabular}
}

\caption{
Omission error counts divided by all testing instances (\%) when training with 1\% data.
}
\label{table:absolute_error_counts}
\end{center}
\end{table}

We additionally show Slot Accuracy (SA) on the non-empty testing instances in \tbref{table:sa_not_none}. The result suggests that our model performs better than \citeauthor{li-etal-2021-zero} in 4 out of 5 domains except for the ``Taxi'' domain, which is the most ``none'' dominated one.
\begin{table}[h]
\begin{center}
{
\small
\setlength\tabcolsep{4pt}
\begin{tabular}{lrrrrr}
\toprule
Model
& Attr.
& Hotel
& Rest.
& Taxi
& Train
\\
\midrule
\citeauthor{li-etal-2021-zero}
& 55.34
& 66.37
& 72.17
& \textbf{24.96}
& 84.70
\\
Ours
& \textbf{61.58}
& \textbf{75.63}
& \textbf{78.06}
& 19.74
& \textbf{85.61}
\\
\bottomrule
\end{tabular}
}

\caption{
Slot Accuracy (\%) on non-empty instances when training with 1\% data.
}
\label{table:sa_not_none}
\end{center}
\end{table}

Both observations indicate that even if omission error occupies more relative proportions of the error cases, our model is able to generate a reasonable amount of not-none values for non-empty slots compared with \citeauthor{li-etal-2021-zero} in most domains.
}
\section{Details of Implementation and Experiments}

\subsection{Implementation Details}

We apply different learning rate optimization for the parameters of each prompt segment. We use separate prompt embeddings for each prompt segment, meaning even if the same token appears in the prefix and question segments during initialization, it maps to different prompt embeddings for a larger optimization space. We use GPT2-medium with 1024 hidden states as our default model.
We use the BPE tokenizer to convert the input sequence to tokens. We set the maximum sequence length to 1024. If the input sequence exceeds the maximum length, we cut the earlier part of the dialogue history while keeping the full other partition.  Only the exact match between the generated sequence and the ground-truth slot value counts as a correct prediction. We use greedy decoding to generate the predicted sequence, and we stop the generation either when \texttt{<|endoftext|>} token is generated or the output length reaches 20. We choose the best epoch by monitoring the JGA of the development set.

Our entire codebase is implemented in PyTorch.\footnote{https://pytorch.org/}
The implementations of the transformer-based models are extended from the Huggingface\footnote{https://github.com/huggingface/transformers}~codebase~\cite{wolf-etal-2020-transformers}.

\subsection{Number of Task Prompt Tokens}
\label{app:best_prompt_num}

We explore various values for the number of task prompt tokens used by the domain, slot and type prompt segments, and we show the hyper-parameters that lead to the best performance in \tbref{table:best_prompt_num}. We observe that the domains with fewer unique slots (such as the attraction domain with just 3 unique slots) need much fewer prompt tokens than the domains with more unique slots (such as hotel and restaurant with 10 and 7 unique slots respectively). The more special prompt tokens needed, the more the parameter numbers are.
\begin{table}[h]
\begin{center}
{
\setlength\tabcolsep{2pt}
\begin{tabular}{lrrrrr}
\toprule
Model
& Attr.
& Hotel
& Rest.
& Taxi
& Train
\\
\midrule
w/ reiteration
& 5
& 20
& 20
& 10
& 10
\\
w/o reiteration
& 5
& 5
& 20
& 5
& 20
\\
\bottomrule
\end{tabular}
}

\caption{
Best prompt numbers for each domain.
}
\label{table:best_prompt_num}
\end{center}
\end{table}

\subsection{Experiment Details}
\label{app:experiment_setting_details}
We report the averaged result for three runs with different random seeds for each experiment. In the ablation study shown in \tbref{table:ablation}, for lines 1-3, we directly remove the corresponding prompt segment from the input sequence; for lines 4-5, we keep the prefix and question text in the input sequence but use token embeddings rather than prompt embeddings to get initial token representation. 
In the prompt token semantic analysis in \tbref{table:cloest_tokens}, we select the most changed prompt tokens by calculating the L2 norm of the difference of the learned and initialized prompt token embeddings.

All the models in this work are trained on a single Nvidia A6000 GPU on a Ubuntu 20.04.2 operating system.
We show the hyperparameter search range and best hyperparameter setting in \tbref{table:hyperparam}.
\begin{table*}[h]
\begin{center}
{
\small
\setlength\tabcolsep{2pt}
\begin{tabular}{lll}
\toprule
Hyperparameter & Search Range & Best \\
\midrule
Number of task prompt tokens & 1, 2, 3, 5, 10, 15, 20, 25, 30 & See \tbref{table:best_prompt_num}\\
Prompt initialization & random, token embedding of task name & token embedding of task name\\
Batch size & 1, 2, 3, 4 & 4\\
Learning rate & 1e-2, 5e-3, 1e-3, 5e-4, 1e-4, 5e-5, 1e-5 & 1e-3 \\
Decoding method & beam search, greedy & greedy \\
Surface form for empty slot & ``none'', ``not mentioned'', ``'' & ``none'' \\
Optimizer & Adam, Lamb \cite{You2020Large} & Adam \\
Early stopping patience epochs & & 8 \\
Learning rate scheduler & & ReduceLROnPlateau with 5 patience epochs\\
Max epochs & & 100 \\
\bottomrule
\end{tabular}
}
\caption{Hyperparameter search range and the best setting.}
\label{table:hyperparam}
\end{center}
\end{table*}

\subsection{Details of the Baseline Models}
\label{app:baseline_models_details}
We produce the result of \citeauthor{li-etal-2021-zero} and \citeauthor{lee-etal-2021-dialogue} with our own reimplementation with our experimental setting and obtain the results of T5DST \cite{lin-etal-2021-leveraging} by running their codebase with our setting. We verify the correctness of our reproduction and we are able to reproduce the performance claimed in their papers under their settings. We report performance for TRADE and DSTQA from their papers. 
For \citeauthor{li-etal-2021-zero}, we use GPT2-medium as the backbone PLM and do not use DSTC8 for transfer learning as it would introduce additional data resources and make the comparison not fair.
For T5DST, we use the best setting concluded by the authors that includes slot type information in the input sequence. We use T5-small with 60M parameters which has 6 encoder-decoder layers and the hidden size of 512 as the backbone PLM. For \citeauthor{lee-etal-2021-dialogue}, we use T5-small as the backbone, we include slot description from MultiWoZ 2.2, possible slot values from dialogue ontology and no domain description in the natural language augmented prompt.

\begin{table*}[ht]
\begin{center}
\resizebox{\linewidth}{!}{
{
\setlength\tabcolsep{2pt}
\begin{tabular}{llll}
\toprule
Domain
& Slot
& Question
& Description
\\
\midrule
\multirow{3}{*}{Attraction} & area 
& In what area is the user looking for an attraction?
& area to search for attractions
\\
& name
& What is the name of the attraction the user prefers?
& name of the attraction
\\ 
& type
& What type of attraction does the user prefer?
& type of the attraction
\\ \midrule
\multirow{10}{*}{Hotel} & area 
& In what area is the user looking for a hotel?
& area or place of the hotel
\\
& book day 
& The user is looking for a hotel starting what day of the week?
& day of the hotel booking
\\
& book people 
& How many people does the user need a hotel booking for?
& number of people for the hotel booking
\\
& book stay 
& How many days does the user prefer to stay at a hotel?
& length of stay at the hotel
\\
& internet 
& Does the user want internet in their hotel?
& whether the hotel has internet
\\
& name 
& What is the name of the hotel the user prefers?
& name of the hotel
\\
& parking 
& Does the user need the hotel to have parking?
& whether the hotel has parking
\\
& price range 
& What is the price range of the hotel the user prefers?
& price budget of the hotel
\\
& stars 
& The user prefers a hotel with what star rating?
& star rating of the hotel
\\
& type 
& What type of hotel does the user prefer?
& what is the type of the hotel
\\ \midrule
\multirow{7}{*}{Restaurant} & area
& In what area is the user looking for a restaurant?
& area or place of the restaurant
\\
& book day 
& The user is looking for a restaurant for what day of the week?
& day of the restaurant booking
\\
& book people 
& How many people does the user need a restaurant booking for?
& how many people for the restaurant reservation
\\
& book time 
& What time does the user want to book a restaurant?
& time of the restaurant booking
\\
& food
& The user prefers a restaurant serving what type of food?
& the cuisine of the restaurant you are looking for
\\
& name
& What is the name of the restaurant the user prefers?
& name of the restaurant
\\
& price range
& What is the price range of the restaurant the user prefers?
& price budget for the restaurant
\\ \midrule
\multirow{4}{*}{Taxi} & arrive by
& What time does the user want to arrive by taxi?
& arrival time of taxi
\\
& departure
& Where does the user want the taxi to pick them up?
& departure location of taxi
\\
& destination
& Where does the user want to go by taxi?
& destination of taxi
\\
& leave at
& What time does the user want the taxi to pick them up?
& leaving time of taxi
\\ \midrule
\multirow{6}{*}{Train} & arrive by
& What time does the user want to arrive by train?
& arrival time of the train
\\
& book people
& How many people does the user need train bookings for?
& how many train tickets you need
\\
& day
& What day does the user want to take the train?
& day of the train
\\
& departure
& Where does the user want to leave from by train?
& departure location of the train
\\
& destination
& Where does the user want to go by train?
& destination of the train
\\
& leave at
& What time does the user want the train to leave?
& leaving time for the train
\\
\bottomrule
\end{tabular}
}
}

\caption{
Question and description used in the input sequence for each slot
}
\label{table:app_question}
\end{center}
\end{table*}

\begin{table*}[ht]
\begin{center}
{
\small
\setlength\tabcolsep{2pt}
\begin{tabular}{llll}
\toprule
Domain & Prompt segment
& Prompt token
& Closest tokens
\\
\midrule
\multirow{10}{*}[-11.8pt]{Attraction} & \multirow{2}{*}{Domain}
& <domain\_attraction\_4> & raction; ractions; racted; ract; ractive
\\
& 
& <domain\_attraction\_0> & att; Att; ATT; atts; atten
\\
\cmidrule{2-4}
& \multirow{2}{*}{Slot}
& <slot\_name\_2>& name; Name; names; NAME; named
\\
&
& <slot\_type\_4>& type; types; Type; style; TYPE
\\
\cmidrule{2-4}
& \multirow{2}{*}{Type}
& <type\_open\_3> & open; Open; opened; opens; opening
\\
&
& <type\_categorical\_3> & ateg; orical; orically; ategy; ategic
\\
\cmidrule{2-4}
& \multirow{2}{*}{Prefix}
& special & special; Special; statistical; SPECIAL; remarkable
\\
&
& site & site; sites; website; Site; webpage
\\
\cmidrule{2-4}
& \multirow{2}{*}{Question}
& Q & answer; Answer; answered;  answers; Q
\\
&
& attraction & attraction; attractions; fascination; attractiveness; attracted

\\
\midrule
\multirow{10}{*}[-11.8pt]{Hotel} & \multirow{2}{*}{Domain}
& <domain\_hotel\_11> & cogn; izoph;  nostalg;  contrad;  Alas
\\
&
& <domain\_hotel\_19> & enment; Alas; ishy;  ridic; minent
\\
\cmidrule{2-4}
& \multirow{2}{*}{Slot}
& <slot\_parking\_13> & Pear; Aqua; Icar; Mermaid; Omega
\\
&
& <slot\_internet\_17> & internet; Wi; Internet;  WiFi; VPN
\\
\cmidrule{2-4}
& \multirow{2}{*}{Type}
& <type\_number\_14> & regex; NUM; abulary; pmwiki; printf
\\
&
& <type\_open\_3> & open;  Dar;  Ezek; Zur; Citiz
\\
\cmidrule{2-4}
& \multirow{2}{*}{Prefix}
& ). & .).; ).; ].; .\}; .</; .]; .); .'; \}.; .)
\\
&
& yes & yes; Yes; YES; yeah; ye
\\
\cmidrule{2-4}
& \multirow{2}{*}{Question}
& hotel & cannabis;  sushi;  Tinder;  whiskey;  booze
\\
&
& days & days;  hours;  consequences;  minutes;  Days
\\
\midrule
\multirow{10}{*}[-11.8pt]{Restaurant} & \multirow{2}{*}{Domain}
& <domain\_restaurant\_7> & rest; Rest; urnal; restrial; restling
\\
&
& <domain\_restaurant\_16> & rest; Rest;  Rest;  Text;  Funds
\\
\cmidrule{2-4}
& \multirow{2}{*}{Slot}
& <slot\_name\_3> & name; Name; names; named; NAME
\\
&
& <slot\_area\_14> & area; Area; But; At; <|endoftext|>
\\
\cmidrule{2-4}
& \multirow{2}{*}{Type}
& <type\_categorical\_4> & orical;  brut; oric; ateg; 
\\
&
& <type\_day\_15> & day; Day; week; DAY; month
\\
\cmidrule{2-4}
& \multirow{2}{*}{Prefix}
& west & west; West; southwest; Southwest;  northwest
\\
&
& booking & booking;  reservation; insult; reverence; audition
\\
\cmidrule{2-4}
& \multirow{2}{*}{Question}
& name & exting; bookstore; describ; mascara; homepage
\\
&
& restaurant & Deadpool; Bitcoin; Veg; steak; Hollywood
\\
\midrule
\multirow{10}{*}[-11.8pt]{Taxi} & \multirow{2}{*}{Domain}
& <domain\_taxi\_6> & i; a; o; I; in
\\
&
& <domain\_taxi\_9> & tax; Tax; taxes; Taxes; taxed
\\
\cmidrule{2-4}
& \multirow{2}{*}{Slot}
& <slot\_destination\_8> & dest; Dest; destination; Destination
\\
&
& <slot\_departure\_4> & ure; ures; URE; ured; uring
\\
\cmidrule{2-4}
& \multirow{2}{*}{Type}
& <type\_time\_0> & time; Time; TIME; timer; year
\\
&
& <type\_open\_6> & open; Open; opens; opening; closed
\\
\cmidrule{2-4}
& \multirow{2}{*}{Prefix}
& open & open; Open; opened; opens; OPEN
\\
&
& rival & rival; rivals; quickShip
\\
\cmidrule{2-4}
& \multirow{2}{*}{Question}
& taxi &  taxi; taxis; Taxi; cab; Uber
\\
&
& does & does;  is;  doesn;  has;  isn
\\
\midrule
\multirow{10}{*}[-11.8pt]{Train} & \multirow{2}{*}{Domain}
& <domain\_train\_8> & train; Train; genre; disciplinary; trained
\\
&
& <domain\_train\_7> & train; Train; trainers; trains; trained
\\
\cmidrule{2-4}
& \multirow{2}{*}{Slot}
& <slot\_destination\_8> & dest;  kosher;  GMO; mill;  JFK; okemon
\\
&
& <slot\_arriveby\_3> & by; By; BY; While; from
\\
\cmidrule{2-4}
& \multirow{2}{*}{Type}
& <type\_time\_0> & time; groupon; times; TIME; wasteful
\\
&
& <type\_day\_6> & day; Day; week; month; year
\\
\cmidrule{2-4}
& \multirow{2}{*}{Prefix}
& dest & dest;  goto; inion; externalTo; Destination
\\
&
& location & location; locations; geographic; geography; geographical
\\
\cmidrule{2-4}
& \multirow{2}{*}{Question}
& from & graphs;  ancestry;  statistics;  backstory;  stats
\\
&
& train & train; plane; railway; subway; highway
\\
\bottomrule
\end{tabular}
}

\caption{
Closest tokens for the learned prompt tokens.
}
\label{table:app_cloest_tokens}
\end{center}
\end{table*}

\end{document}